\documentclass[letterpaper]{article}

\usepackage[preprint]{neurips_2022}

\usepackage[utf8]{inputenc} 
\usepackage[T1]{fontenc}    
\usepackage{hyperref}       
\usepackage{url}            
\usepackage{booktabs}       
\usepackage{amsfonts}       
\usepackage{nicefrac}       
\usepackage{microtype}      
\usepackage{xcolor}     
\usepackage{graphicx}
\usepackage{amsmath}
\usepackage{amsthm}
\usepackage{booktabs}
\usepackage{algorithm}
\usepackage{algorithmic}    

\newtheorem{theorem}{Theorem}

\newtheorem{definition}{Definition}
\newcommand{\cu}[1]{\boldsymbol{#1}}
\usepackage{abstract}

\title{Parameter Convex Neural Networks}

\author{%
	Jingcheng Zhou$^{1,2}$\And
	Wei Wei$^{1,2,3,4,}$\And
	Xing Li$^{1,2}$\And
	Bowen Pang$^{1,2}$\And
	Zhiming Zheng$^{1,2,3,4}$ \\
	$^1$School of Mathematical Sciences, Beihang University, Beijing, China\\
	$^2$Key Laboratory of Mathematics Informatics Behavioral Semantics, Ministry of Education, China\\
	$^3$Institute of Artificial Intelligence, Beihang University, Beijing, China\\
	$^4$Peng Cheng Laboratory, Shenzhen, Guangdong, China\\
	\texttt{ \{17374347,weiw,sy1809120,pangbw\}}@buaa.edu.cn\\
	\texttt{zzheng@pku.edu.cn}
}

\begin{document}
	\maketitle

	\begin{abstract}
		Deep learning utilizing deep neural networks (DNNs) has achieved a lot of success recently in many important areas such as computer vision, natural language processing, and recommendation systems. The lack of convexity for DNNs has been seen as a major disadvantage of many optimization methods, such as stochastic gradient descent, which greatly reduces the genelization of neural network applications. We realize that the convexity make sense in the neural network and propose the exponential multilayer neural network (EMLP), a class of parameter convex neural network (PCNN) which is convex with regard to the parameters of the neural network under some conditions that can be realized. Besides, we propose the convexity metric for the two-layer EGCN and test the accuracy when the  convexity metric changes. For late experiments, we use the same architecture to make the exponential graph convolutional network (EGCN) and do the experiment on the graph classificaion dataset in which our model EGCN performs better than the graph convolutional network (GCN) and the graph attention network (GAT).
	\end{abstract}

	\section{Introduction}
	Deep neural networks have received flood of attention
	on account of unbeatable success in many fields, such as computer
	vision \cite{krizhevsky2012imagenet}, natural language processing \cite{hinton2012deep} and recommendation system \cite{guo2017deepfm}. Despite good performance by some first-order gradient based optimization algorithms such as stochastic gradient descent (SGD), it is still theoretically incomprehensible that how and why such algorithms can train deep neural networks (DNNs) successfully. Optimization algorithms such as stochastic gradient descent (SGD) cause many problems that the optimization algorithms are not sure to converge to a stable point and the global minimum points are hard to get. However, even though the optimization algorithms can converge to a stable point, it's still  hard to decide whether the stable point is a global minimum point even local minimum point.
	
	The global minimum point is hard to solve in nonconvex problems and deep learing is just a representative nonconvex problem that the object function is very complicated and nonconvex. The loss function has many critical points to make the problem more difficult and there are many works about how a critical point can be a global minimum point.
	For deep linear neural networks, a sufficient and necessary condition that the critical point of the objective function is the global minimum point is proposed by \cite{yun2017global}. Besides, \cite{haeffele2017global} also shows that the global minimum point can just be the local minimum point under sufficient conditions, and when they make the network output and regularization to be positive homogeneous functions of network parameters with the regularization design  used to control the size of the network, the local descent method can achieve the global minimum under any initialization conditions. \cite{boob2017theoretical} shows that for a broad class of differentiable activation functions, the first-order optimal solution can be global minimum point as long as the hidden layer is non-singular. \cite{nguyen2017loss} shows that (almost) so, in fact almost all local minimun points are global optimal minimun points, and that the fully connected network with square losses and the analytic activation function assume that the number of hidden quantitative units in a layer of networks is greater than the training fraction, and that the network structure of this layer is pyramidal. The most popular algorithm for deep learning optimization is stochastic gradient descent and there are lots of works about the global minimum point when utilizing SGD under different limited conditions. \cite{li2017convergence} shows that SGD converges to the global minimum point with a polynomial number of steps and is initialized with the standard deviation $O(1/d)$ when the input follows from a Gaussian distribution.
	\cite{kawaguchi2019gradient} proves theoretically that gradient descent method can find the global minimum point of non-convex optimization for all layers of nonlinear deep neural networks of various sizes.
	Unlike gradient descent and quasi-Newton methods, other works put forward some different algorithms that may get the solution which is the global minimum point. \cite{dauphin2014identifying} proposes a new second-order optimization method, the saddle-free Newton method, which can avoid high dimensional saddle points quickly. Also, \cite{wu2019global} proposes an adaptive gradient algorithm and proves that for two-layer over-parameterized neural networks, if the width of the net is large enough (polynomially), the algorithm converges to the global minimum point in polynomial time. Besides, some unique works get the global minimum point in unexpected way. Gradient descent method achieves zero loss in the train dataset in polynomial time for a deep overparameterized neural network \cite{du2019gradient} with residual connections (ResNet). \cite{sun2020global} points out that there may be suboptimal local minimum points in generalized neural networks under certain hypothetical conditions, and discuss some strict results of geometric properties of generalized neural networks, such as no bad basin, and elimination of suboptimal local minimum points and reduction of paths to infinity.
	
	Convergence is also a popular hotspot for training the deep neural networks. When the initialization is correct in the multiparticle limit, the gradient flow, though non-convex, converges to a global minimum point \cite{chizat2018global}. For shallow neural networks with $m$ hidden nodes activated by ReLU and training data, \cite{du2018gradient} proves that as long as $m$ is large enough and there are no two parallel inputs, the randomly initialized gradient descent function converges to the global optimal minimum solution at a linear convergence rate. \cite{wang2018exponential} proves that the convergence rate of deep neural network approximations is exponential for low dimensional analytic functions. \cite{allen2019convergence} shows that simple algorithms, such as stochastic gradient descent (SGD), can find global minimum point on training datasets in polynomial time, provided that the input does not degenerate and the network is over-parameterized. \cite{sun2020global} briefly discusses some convergence results and their relationship with the neural network landscape results. \cite{zhou2021local} shows that as long as the loss is less than a threshold value, in an over-parameterized double-layer neural network, all the student neurons will converge to one of the teacher neurons and the loss will approach zero.
	
	In this paper, we don't analyse the global minimum point  and convergence for deep neural networks. Instead, we remould the deep neural networks and propose the Parameter Convex Neural Network (PCNN) which is convex to the parameters. We first transform the multilayer neural network (MLP) to exponential multilayer neural network (EMLP) and use the same technique to make the exponential graph convolutional network (EGCN) converted from the GCN. As far as we know, this is the fisrt time that the convex neural network to parameters but not input is proposed. Besides, we experiment our model on classification assignment and get excellent results.

	\section{Models}

	\subsection{Notations, Definitions and Theorems}
	Respectively, we denote the matrices, vectors and scalars as uppercase bold letters (such as $\cu A$), lowercase bold letters (such as $\cu a$) and lowercase letters (such as $a$). To denote a vector or
	matrix of zeros, we use $\cu 0$, where the size is acquired from the context. In addition, We list some notations for the future use.\\
	$\cu A\odot\cu B$\qquad Hadamard product\\
	$\cu A\otimes\cu B$\qquad Kronecker product\\
	$\cu A^n$\qquad $n$ times Hadamard product of $A$\\
	diag($\cu x$)\qquad turn a vector into a diagonal matrix\\
	$vec(\cu W)$\qquad  vec operator transforms a matrix into a vector by stacking the
	columns of the matrix one underneath the other\\
	$a^{\cu W}$\qquad exponential operator transforms a matrix into another matrix with the same size and transform every element $\cu W_{ij}$ into $a^{\cu W_{ij}}$\\
	$log_a(\cu W)$\qquad logarithm operator transforms a matrix into another matrix with the same size and transform every element $\cu W_{ij}$ into $log_a(\cu W_{ij})$\\
	$\cu x>(\ge) \cu 0$\qquad every element $\cu x_i>(\ge)0$\\
	$\cu A\succ(\succeq)0$ \qquad  $\cu A$ is a positive (semi)definite matrix\\
	$\cu A\prec(\preceq)0$ \qquad  $\cu A$ is a negative (semi)definite matrix\\
	
	To make a class of the PCNN, we first  recall the second-order conditions of convex functions. We now assume that $f$ is twice differentiable, that is, its Hessian or second derivative $\bigtriangledown^2 f$  exists at each point in dom $f$, which is open. Then $f$ is convex if and only if dom $f$ is convex and its Hessian is positive semidefinite: \[\forall\cu x\in dom f, \bigtriangledown^2 f(\cu x)\succeq 0.\]
	Because the parameters of the neural network are a series of matrices, we need first to define the Hessian of functions when independent variables are matrices. If $f(\cu X)$ is a scalar function in which $X$ is a matrix, then the Hessian can be simply defined:
	\[
	H=f_{vec(\cu X)(vec(\cu X))^T}
	\]
	which can be equivalent to the Hessian of $f$ to $vec(\cu X)$. This method is very hard to calculate,  so we calculate the Hessian with another method in this paper:
	\[
	d^2f=(d \,vec(\cu X))^T\cu B \,d\,vec(\cu X)
	\]
	so that \[H=\dfrac{1}{2}(\cu B+\cu B^T).\]
	For more information with the Hessian, you can see the book \cite{magnus2019matrix} to learn more accurate definition of the Hessian and more methods to calculate.
	We now define the convex matrix funtion of matrices.
	\begin{definition}[convex matrix funtion]
		Given a matrix function $\cu{F}(\cu x)$ (vector function $\cu f(
		\cu x)$ or scalar function $f(\cu x)$) in whcih $\cu x$ is a scalar, vector, matrix or their conbinations, we say $\cu{F}(\cu x)$ is (strictly) convex corresponding to $\cu x$ if and only if  each element of  $\cu{F}(\cu x)$  is (strictly) convex corresponding to $\cu \theta$ which is a column vector denotation of $\cu x$.
	\end{definition}
	This definition means that no matter what $\cu x$ is, we just consider the $\cu x$ as a column vector to judge the convex function.
	Then we define the PCNN and ICNN in details.
	\begin{definition}[PCNN,ICNN]
		$\cu F(\cu x,\cu\theta)$ is a neural network, in which $\cu x$ is input and $\cu\theta $ represents parameters. We say $\cu F(\cu x,\cu\theta)$ is a Parameter Convex Neural Network (PCNN) if and only if $\cu F(\cu x,\cu\theta)$ is convex with regard to $\cu \theta$. Say $\cu F(\cu x,\cu\theta)$ is the Input Convex Neural Network (ICNN) if and only if $\cu F(\cu x,\cu\theta)$ is convex with regard to $\cu x$.
	\end{definition}
	The late proof will calculate the Hessian when the argument $\cu W$ is a matrix. Just as the definition of the PCNN and the ICNN, we calculate the gradient and Hessian for $vec(\cu W)$.
	
	\subsection{Multilayer Neural Network}
	A simple layer of the multilayer neural network (MLP) $f(\cu W, \cu x)$ can be defined as:
	\begin{equation}
		\cu x^{(n+1)}=\sigma({\cu W_{n}}\cu x^{(n)})\label{1}
	\end{equation}
	in which $\cu W_{n}$ is the parameters of the $n$-th layer, $\cu x^{(n)}$ is the $n-$th layer and $\sigma$ is the activation function with $\cu x^{(0)}=\cu x$ is the input and $\cu W$ is all parameters of $f$. We define the MLP $f_{n}(\cu W, \cu x)$ of $n$ layers as:
	\begin{equation}
		f_{n}(\cu W, \cu x)=\cu x^{(n)}={\cu W_{n-1}}\cu x^{(n-1)}
	\end{equation}We just calculate the Hessian of the MLP (input and output are scalars) of two layers:
	\begin{equation}
		f(\cu W, \cu x)=\cu W_{1}^T\sigma(\cu W_{0} x).
	\end{equation}
	in which $\cu W_{1}$ and $\cu W_{0}$ are column vectors.
	We just list the result of the computing without calculation details:
	\begin{align}
		f_{\cu W_{1}\cu W_{1}}&=0\\
		f_{\cu W_{1}\cu W_{0}}&= x\otimes diag(\sigma^{'}(\cu W_{0} x))\\
		f_{\cu W_{0}\cu W_{0}}&=x^2diag(\sigma^{''}(\cu W_{0}x)\odot\cu W_{1}).
	\end{align}
	Then we can get the Hessian: 
	\begin{equation}
		H(f)=\begin{bmatrix}
			f_{\cu W_{1}\cu W_{1}}&f_{\cu W_{1}\cu W_{0}}\\
			f^T_{\cu W_{1}\cu W_{0}}&f_{\cu W_{0}\cu W_{0}}
		\end{bmatrix}
	\end{equation}
	in which we just use the symbol $f_{\cu W\cu W}$ to replace $f_{vec(\cu W)(vec(\cu W))^T}$ for simplicity.
	For the computing details, you can learn methods from the book \cite{magnus2019matrix}.
	From the Hessian, we can know that to make the Hessian positive definite, we need to transform the Hessian to make $f_{\cu W_{1}\cu W_{1}}\succeq 0$.
	\subsection{Exponential Multilayer Neural Network}
	We think a lot of methods to make $f_{\cu W\cu W}\succeq 0$, such as transforming $\cu W$ to $\cu W^2$ or $a^{\cu W}$.
	The model that we create is the Exponential Multilayer Neural Network (EMLP). Just as its name implies, EMLP uses exponential layer. The exponential layer is defined as:
	\begin{equation}
		\cu x^{(n+1)}=\sigma(a^{\cu W_{n}}\cu x^{(n)})\label{2}.
	\end{equation}
	in which $\cu x^{(n)}$ is the $n$-th layer of the EMLP and $\sigma$ is the activation function with the  exponential hyperparameter $a$. We define the EMLP $f_{a,n}(\cu W, \cu x)$ of $n$ layers as:
	\begin{equation}
		f_{a,n}(\cu W, \cu x)=\cu x^{(n)}=a^{\cu W_{n-1}}\cu x^{(n-1)}
	\end{equation}
	in which $\cu x^{(n-1)}$ can be get from the iteration equation \ref{2} and $\cu x^{(0)}=\cu x$ with the exponential hyperparameter $a$. The last layer of the EMLP $f_a(\cu W, \cu x)$ is with no activation function which is natural for regression assignments. For classifiction assignments, we take softmax funcion as one part of the cross entropy function.
	
	We first prove the convexity of a simply condition of the EMLP. All the proofs are put in supplementary material.
	\begin{theorem}
		Let scalar $ x> 0$, $\sigma( x)> 0$, $\sigma^{'}( x)> 0$, $\sigma^{''}( x)> 0$, $a>0$ and $a \ne 1$. Let \[
		f(\cu W)=a^{\cu W_2^T}\sigma(a^{\cu W_1} x)
		\] in which $\cu W_1$ and $\cu W_2$ are column vectors and $\cu W$ represents parameters $\cu W_1$ and $\cu W_2$. Then  $f(\cu W)$ is convex corresponding to $\cu W$ if and only if \[\forall x>0, x\sigma^{''}(x)\sigma(x)+\sigma(x)\sigma^{'}(x)\ge x\sigma^{'2}(x).\]
	\end{theorem}
	
	Theorem 1 just presents the simplest condition of the EMLP. We will give a more general of the EMLP int the following theorems.
	\begin{theorem}
		Suppose $ \cu x> \cu 0$, $\sigma( \cu x)> \cu 0$, $\sigma^{'}(\cu x)> \cu 0$, $\sigma^{''}( \cu x)> \cu 0$, $a>0$ and $a \ne 1$. And the activation function meets with: \[\forall x>0, x\sigma^{''}(x)\sigma(x)+\sigma(x)\sigma^{'}(x)\ge x\sigma^{'2}(x).\] Then $f_{a,2}(\cu W,\cu x)$ is convex corresponding to $\cu W$ or we say $f_{a,2}(\cu W,\cu x)$ is a PCNN.
	\end{theorem}
	
	Theorem 2 shows that a two-layer EMLP is a PCNN under some conditions. We will show the conditions of $n$-layer EMLP which is more complicated.
	\begin{theorem}
		Suppose $ \cu x> \cu 0$, $\sigma( \cu x)> \cu 0$, $\sigma^{'}(\cu x)> \cu 0$, $\sigma^{''}( \cu x)> \cu 0$, $a>0$ and $a \ne 1$. And the activation function meets with: \[\forall x>0, \sigma^{''}(x)\sigma(x)\ge \sigma^{'2}(x).\] Then \[f(\cu\theta)=a^{\cu \theta_1^T}\sigma(\cu\theta_2)\] is convex corresponding to $\cu\theta$ in which \[\cu\theta=[\cu\theta_1^T,\cu\theta_2^T]\] and $\cu\theta_1$ and $\cu\theta_2$ are column vectors.
	\end{theorem}
	
	\begin{theorem}
		Suppose $ \cu x> \cu 0$, $\sigma( \cu x)> \cu 0$, $\sigma^{'}(\cu x)> \cu 0$, $\sigma^{''}( \cu x)> \cu 0$, $a>0$ and $a \ne 1$. And the activation function meets with: \[\forall x>0, \sigma^{''}(x)\sigma(x)\ge \sigma^{'2}(x).\] Then $f_{a,n}(\cu W,\cu x)$ is a PCNN.
	\end{theorem}
	
	\subsection{Graph Convolutional Neural Network}
	A two-layer GCN model takes the simple form:
	\[
	\cu f(\cu X,\cu W)=\cu A\sigma(\cu A\cu X\cu W_{0})\cu W_{1}
	\]
	in which $\cu A$ is the adjacency matrix.
	Our model Exponential Graph Convolutional Network (EGCN) of two layers can be defined as:
	\begin{equation}
		\cu f_{a,2}(\cu X,\cu W)=\cu A\sigma(\cu A\cu Xa^{ \cu W_{0}})a^{ \cu W_{1}}.
	\end{equation}
	We define the EMLP $f_{a,n}(\cu W, \cu x)$ of $n$ layers as:
	\begin{equation}
		\cu f_{a,n}(\cu X, \cu W)=\cu X^{(n)}=\cu A\cu X^{(n-1)}a^{\cu W_{n-1}}.
	\end{equation}
	We first prove the condition of a two-layer EGCN.
	\begin{theorem}
		Suppose $ \cu X> \cu 0$, $\sigma( \cu X)> \cu 0$, $\sigma^{'}(\cu X)> \cu 0$, $\sigma^{''}( \cu X)> \cu 0$, $a>0$ and $a \ne 1$. And the activation function meets with: $\forall x>0, x\sigma^{''}(x)\sigma(x)+\sigma(x)\sigma^{'}(x)\ge x\sigma^{'2}(x)$. Then $\cu f_{a,2}(\cu X,\cu W)$ is a PCNN.
	\end{theorem}
	
	Then, we prove the condition of the EGCN with $n$ layers.
	\begin{theorem}
		Suppose $ \cu X> \cu 0$, $\sigma( \cu X)> \cu 0$, $\sigma^{'}(\cu X)> \cu 0$, $\sigma^{''}( \cu X)> \cu 0$, $a>0$ and $a \ne 1$. And the activation function meets with: \[\forall x>0,\sigma^{''}(x)\sigma(x)\ge \sigma^{'2}(x).\] Then $\cu f_{a,n}(\cu X,\cu W)$ is a PCNN.
	\end{theorem}
	
	\section{Convexity of parameter convex neural network}
	To meet the limited condition of parameter convex neural network, we need do some preprocesss.
	\subsection{Data preprocess}
	One of the limited conditions is that the input must be greater than zero. All the datasets that we utilize are no less than zero so that we need transform the input to meet with the condition. The method we put forward is:
	\begin{equation}
		\begin{cases}
			\cu X_{ij}\to b^{\cu X_{ij}},\,\cu X_{ij}>0\\
			\cu X_{ij}\to \epsilon,\,\cu X_{ij}=0\\
		\end{cases}
	\end{equation}
	in which $b$ is a constant and $\epsilon$ is a small decimal. We set $\epsilon$ to meet with the condition ($\cu X>0$) and $b$ is to make the elements that are greater than zero more bigger than $\epsilon$.
	\subsection{Activation function}
	We have two different conditions for the activation function. The fisrt condition is:
	\begin{align*}
		\forall x>0,\,\sigma(x)>0,\,\sigma^{'}(x)>0,\,\sigma^{''}(x)>0,
		x\sigma^{''}(x)\sigma(x)+\sigma(x)\sigma^{'}(x)\ge x\sigma^{'2}(x).
	\end{align*}
	We find three activation functions meeting with the first condition: \[relu(x)=\max(x,0),softplus(x)=\ln(1+e^x),bend(x)=\dfrac{\sqrt{x^2+1}-1}{2}+x.\]
	We need to notice that the three activation functions plus any nonnegative constant and the combinations of the three activation function meet the first condition.
	
	The second condition is:
	\begin{align*}
		\forall x>0,\,\sigma(x)>0,\,\sigma^{'}(x)>0,\,\sigma^{''}(x)>0,\sigma^{''}(x)\sigma(x)\ge \sigma^{'2}(x).
	\end{align*}
	
	We find no activation functions meeting with the second condition and some other funtions like $e^x$ will make many problems such as the output will be infinite so that we use just do the experiments with the EGCN of two-layers.
	\subsection{Convex optimization}
	
	For classification assign, we have no idea how to transform it to a convex function because the softmax function will make a problem. The softmax function:
	\[
	f(\cu x)=\dfrac{e^{\cu x}}{\cu 1^Te^{\cu x}}
	\]
	is not monotone increasing or monotone decreasing which means
	\[
	\bigtriangledown f(\cu x)\ge \cu0 \,(or \le\cu0)
	\]
	is not satisfactory.
	\subsection{Convexity metric}
	The convexity metric reflects the convexity of the convex funtion and the The second derivative can report the curve flexibility. We define the convexity metric of $f(\cu x)$ is the determinant of the Hessian:
	\[
	\mathcal{C}=|H(f(\cu x))|.
	\]
	From the supplementary materials, we get the Hessian of the two-layer EGCN:
	\begin{equation}
		H(f)=\begin{bmatrix}
			f_{\cu W_{1}\cu W_{1}}&f^T_{\cu W_{2}\cu W_{1}}\\
			f_{\cu W_{2}\cu W_{1}}&f_{\cu W_{2}\cu W_{2}}
		\end{bmatrix}
	\end{equation}
	in which
	\begin{align}
		f_{\cu W_2\cu W_2}&=\ln^2 a\, diag(\sigma(a^{\cu W_1} x)\odot a^{\cu W_2})\\
		f_{\cu W_2\cu W_1}&=x\ln^2 a\,diag(\sigma^{'}(a^{\cu W_1} x)\odot a^{\cu W_2+\cu W_1})\\
		f_{\cu W_1\cu W_1}&=x^2\ln^2 a\,diag(\sigma^{''}(a^{\cu W_1} x)\odot a^{\cu W_2+2\cu W_1})\\&+x\ln^2 a\,diag(\sigma^{'}(a^{\cu W_1} x)\odot a^{\cu W_2+\cu W_1}).
	\end{align}
	Because of the properties of partitioned matrices, we can get the convexity metirc:
	\[
	\mathcal{C}=|f_{\cu W_2\cu W_2}f_{\cu W_1\cu W_1}-f_{\cu W_2\cu W_1}f_{\cu W_2\cu W_1}|.
	\]
	This definition is very complicated and hard to tell which convexity metric is bigger so that we get the simple definition:
	\[
	\mathcal{C}=\int_0^1 x\sigma^{''}(x)\sigma(x)+\sigma(x)\sigma^{'}(x)-x\sigma^{'2}(x)dx
	\]
	\section{Experiments}
	Experiments on  classification problems are led on the three datasets summarized in Table \ref{tab1}. This section summarizes our experimental setup and results.
	
	\begin{table}[H]
		\centering
		\caption{Summary of the datasets used in our experiments.}
		{
			\begin{tabular}{lllll}
				\toprule
				& \bf{Cora} & \bf{CiteSeer} & \bf{PubMed}\\
				\midrule
				\bf{\# Nodes}      &2708 (1 graph) &3327 (1 graph) &19717 (1 graph)  \\
				\bf{\# Edges}      &5429 &4732 &44338      \\
				\bf{\# Features}   &1433 &3703 &500     \\
				\bf{\# Classes}   & 7  & 6     &3      \\
				\bf{\# Training Nodes}   & 140  & 120 &60     \\
				\bf{\# Validation Nodes}   & 500  & 500 &500     \\
				\bf{\# Test Nodes}   & 1000  & 1000 &1000     \\
				\bottomrule
		\end{tabular}}
		
		\label{tab1}
	\end{table}
	
	\subsection{Datasets}
	We utilize three standard citation network datasets: Citeseer, Cora and Pubmed \cite{sen2008collective} which closely follow the transductive experimental setup in \cite{yang2016revisiting}. For training, we only use 20 labels per class, but all feature vectors. The predictive power of the trained models is evaluated on 1000 test nodes, and we use 500 additional nodes for validation purposes (the same ones as used by \cite{kipf2016semi}). The Cora dataset contains 2708 nodes, 5429 edges, 7 classes and 1433 features per node. The Citeseer dataset contains 3327 nodes, 4732 edges, 6 classes and 3703 features per node. The Pubmed dataset contains 19717 nodes, 44338 edges, 3 classes and 500 features per node.
	\subsection{Experiment setup}
	We train a two-layer EGCN described in Section 2.4 and evaluate prediction accuracy on a test set of 1,000 labeled examples. We choose the same dataset splits as in \cite{yang2016revisiting} with an additional validation set of 500 labeled examples. We train Cora and CiteSeer with the hidden layer of 16 nodes for a maximum of 200 epochs(training iterations) and PubMed with the hidden layer of 8 nodes for a maximum of 400 epochs using Adam \cite{kingma2014adam} with a learning rate of 0.01 without dropout and save the model and test when the predicted accuracy rate of the validation set is biggest . We initialize weights using the  default random initialization. For data preprocess, we set $b=2$ for all models, and $\epsilon=0.088$ for Cora and $\epsilon=0.1$ for CiteSeer and PubMed. The activation funtion for all models is softplus ($\ln(1+e^x)$). The exponential hyperparameter $a$ for all models we set is $2$.
	\subsection{Results}
	The results of our comparative evaluation experiments are summarized in Table \ref{tab2}. For the classification tasks, we report the classification accuracy rate on the test nodes of our model, and reuse the metrics already reported in \cite{yang2016revisiting} for state-of-the-art techniques. 
	\begin{table}[H]
		\centering
		\caption{Summary of results in terms of classification accuracies, for Cora, CiteSeer and Pubmed on test datasets.}
		{
			\begin{tabular}{llll}
				\toprule
				\bf{Method}& \bf{Cora} & \bf{CiteSeer} & \bf{PubMed} \\
				\midrule
				MLP      &55.1\% &46.5\% &71.4\%  \\
				Planetoid \cite{yang2016revisiting}      &75.7\% &64.7\% &77.2\%     \\
				MoNet \cite{monti2017geometric}   &81.7\% &- &78.8\%     \\
				Chebyshev \cite{defferrard2016convolutional}   & 81.2\%  & 69.8\%     &74.4\%     \\
				GCN \cite{kipf2016semi}   & 81.5\%  & 70.2\% &78.9\%   \\
				GAT \cite{velivckovic2017graph}   & 83.7\%  & 71.7\% &78.9\%     \\
				EGCN\_bend (ours)   & {83.8}\%  & {71.9}\% &{81.4}\%    \\
				EGCN\_relu (ours)   & \bf{84.6}\%  & \bf{73.6}\% &\bf{82.4}\%    \\
				EGCN\_softplus (ours)   & \bf{84.6}\%  & \bf{73.6}\% &\bf{82.4}\%    \\
				\bottomrule
		\end{tabular}}
		
		\label{tab2}
	\end{table}
	From the Table \ref{tab2}, we can see that our model EGCN is best in those model. We think this is because our model have better generalization ability so that we present the accuracy on validation datasets.
	\begin{table}[H]
		\centering
		\caption{Summary of results in terms of classification accuracies, for Cora, CiteSeer and Pubmed on validation datasets.}
		{
			\begin{tabular}{llll}
				\toprule
				\bf{Method}& \bf{Cora} & \bf{CiteSeer} & \bf{PubMed} \\
				\midrule
				GCN \cite{kipf2016semi}   & 77.8\%  & 70.3\% &78.8\%   \\
				GAT \cite{velivckovic2017graph}   & \bf{82.2}\%  & 71.6\% &79.6\%     \\
				EGCN (ours)   & 81.8\%  & \bf{73.6}\% &\bf{82.8}\%    \\
				\bottomrule
		\end{tabular}}
		\label{tab3}
	\end{table}
	To explore the relationship between $\mathcal{C}$ and the accuracy, we do some other experiments. When $\sigma(x)=relu(x)$, we can get the $\mathcal{C}=0$ which means the activation function $relu(x)$ is a very good function to do the experiment. We just need turn $relu(x)$ to $relu(x)+c$ so that $c$ is just the convexity measure. We do the experiments on the three datasets with $relu(x)+c$ when $c$ changes from 0 to 99 and the result is below. We can see that the accuracy on PubMed dataset decreases as the convexity measure $c$ increases and the accuracy on other two datasets tends to decrease generally as the the convexity measure $c$ increases.
	\begin{figure}[H]
		\centering
		\caption{Accuracies of different convexity measures on three datasets.}
		\includegraphics[scale=0.7]{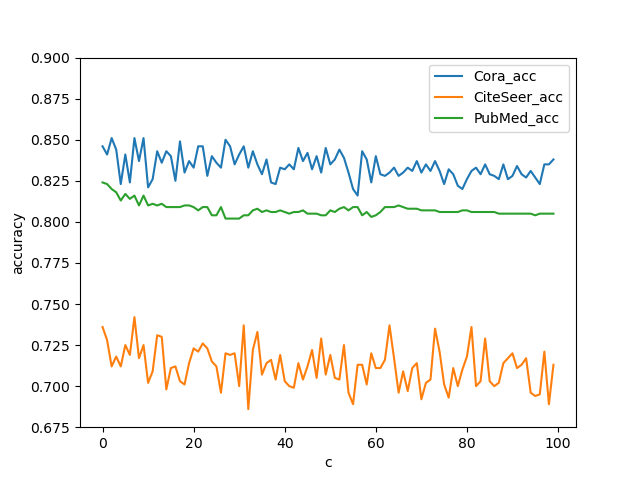}
		
		\label{fig:acc}
	\end{figure}

	\section{Discussion}
	\subsection{Limitations of EMLP and EGCN}
	\paragraph{Approximation capability} We can just think the EMLP as just a kind of the neural network and our model EGCN as a kind of GCN but with all the parameters are larger than zero. The parameters are all limited in the open half space when we train the EGCN which means the approximation capability of EGCN is lower than GCN. The limitaion will cause problems such as the train loss will be much large and the predicted values will deviate much from the real values when the parameters of the trained GCN are mostly less than zero and the parameters of the trained EGCN are mostly be close to negative infinity. Thus, only when the parameters of the trained GCN are mostly nonnegative, the model EGCN can make a satisfactory solution.
	\paragraph{Convexity limits} To get the convexity of the loss function, we give many limits. one of the limits of input is that $\cu x>\cu0$ which is a very strict restriction that in many dataset, input can't meet with it. The activation function that meets with condition is also infrequent. Besides, when training EGCN, it may happen that the value is out of range.
	\subsection{Future work}
	We hope that we can make more kinds of PCNN besides EMLP and EGCN and transform other neural network to be equiped with convexity such as CNN. For EMLP and EGCN, we can think deeply how to reduce of the limited conditions of convexity. At the same time, we can also reflect on how to enhance approximation capability.
	\section{Conclusion}
	In this paper, we transform the neural network to be a convex function of the parameters which is the first convex neural network in our opinion. Besides, we introduce the concept of PCNN  and propose the general technique to make two classes of PCNN: EMLP and EGCN and  do experiments on graph classification datasets which perform better than GCN and GAT.

	\begin{ack}
		This work is supported by the Research and Development Program of China (Grant No. 2018AAA0101100), the National Natural Science Foundation of China (Grant Nos. 62141605, 62050132), the Beijing Natural Science Foundation (Grant Nos. 1192012, Z180005).
	\end{ack}

	\bibliographystyle{named}
	\bibliography{nips2022}

\end{document}